\documentclass[10pt,twocolumn,letterpaper]{article}

\usepackage{cvpr}              
\usepackage{algorithm}
\usepackage{algpseudocode}
\usepackage{array} 
\usepackage{multirow}
\usepackage{graphicx}
\usepackage{array}
\usepackage{multirow} 
\usepackage{colortbl} 
\usepackage{hhline} 

\usepackage[dvipsnames]{xcolor}

\definecolor{cvprblue}{rgb}{0.21,0.49,0.74}
\usepackage[pagebackref,breaklinks,colorlinks,citecolor=cvprblue]{hyperref}

\def\confName{CVPR}
\def\confYear{2024}

\title{GPTDrawer: Enhancing Visual Synthesis through ChatGPT}

\author{
\begin{tabular}{@{}c@{\hspace{2cm}}c@{}}
{\tt\small Kun Li} & {\tt\small Xinwei Chen} \\
{\tt\small University of Illinois Urbana-Champaign} & {\tt\small University of Illinois at Urbana Champaign} \\
{\tt\small Champaign, IL, USA} & {\tt\small Champaign, IL, USA} \\
{\tt\small kunli3@illinois.edu} & {\tt\small xinweic2@illinois.edu} \\[1em]
{\tt\small Tianyou Song} & {\tt\small Hansong Zhang} \\
{\tt\small Columbia University} & {\tt\small University of California San Diego} \\
{\tt\small New York City, NY, USA} & {\tt\small La Jolla, CA, USA} \\
{\tt\small tianyou.song@columbia.edu} & {\tt\small haz064@ucsd.edu} \\[1em]
{\tt\small Wenzhe Zhang} & {\tt\small Qing Shan} \\
{\tt\small University of California San Diego} & {\tt\small Northeastern University} \\
{\tt\small La Jolla, CA, USA} & {\tt\small Seattle, WA, USA} \\
{\tt\small wez007@ucsd.edu} & {\tt\small shan.qi@northeastern.edu}
\end{tabular}
}

\begin{document}
\maketitle

\begin{table*}[h]
\centering
\begin{tabular}{ccc}
  \includegraphics[width=0.3\linewidth]{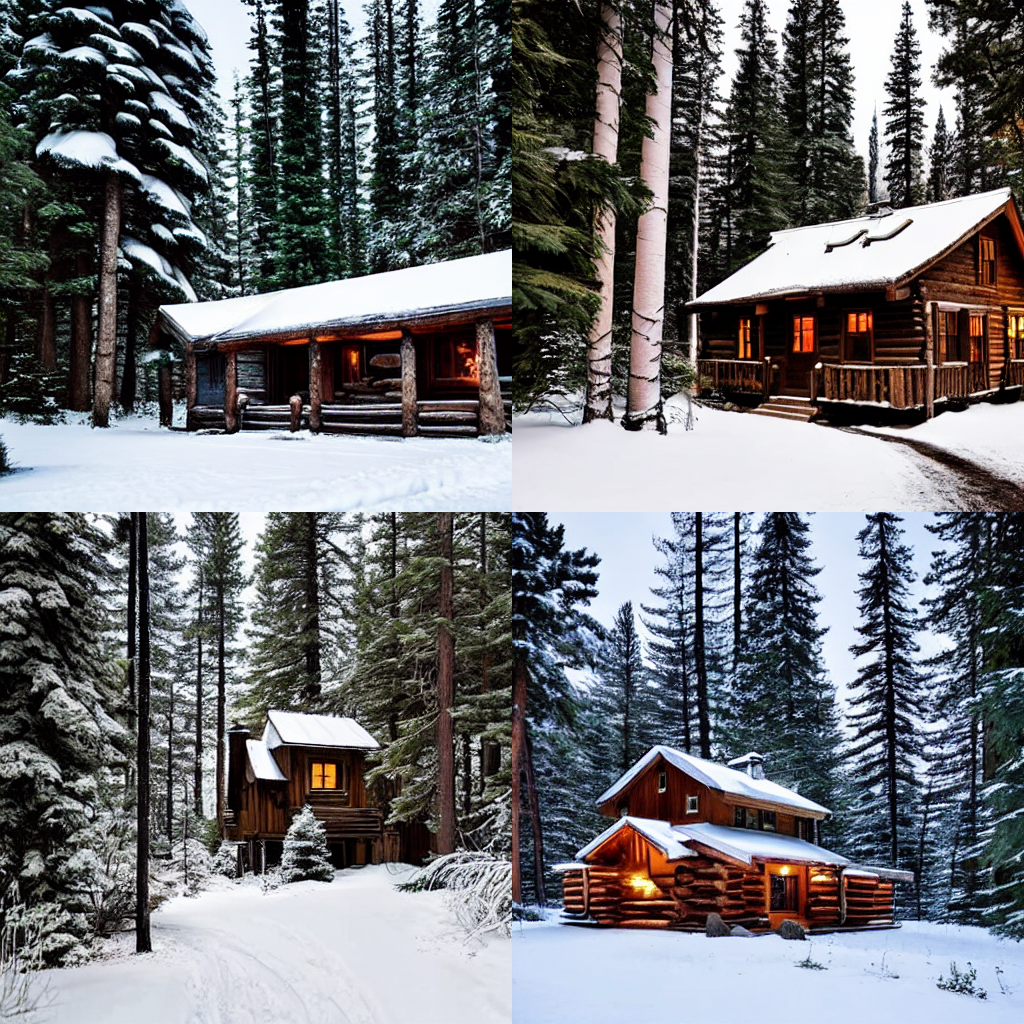} & 
  \includegraphics[width=0.3\linewidth]{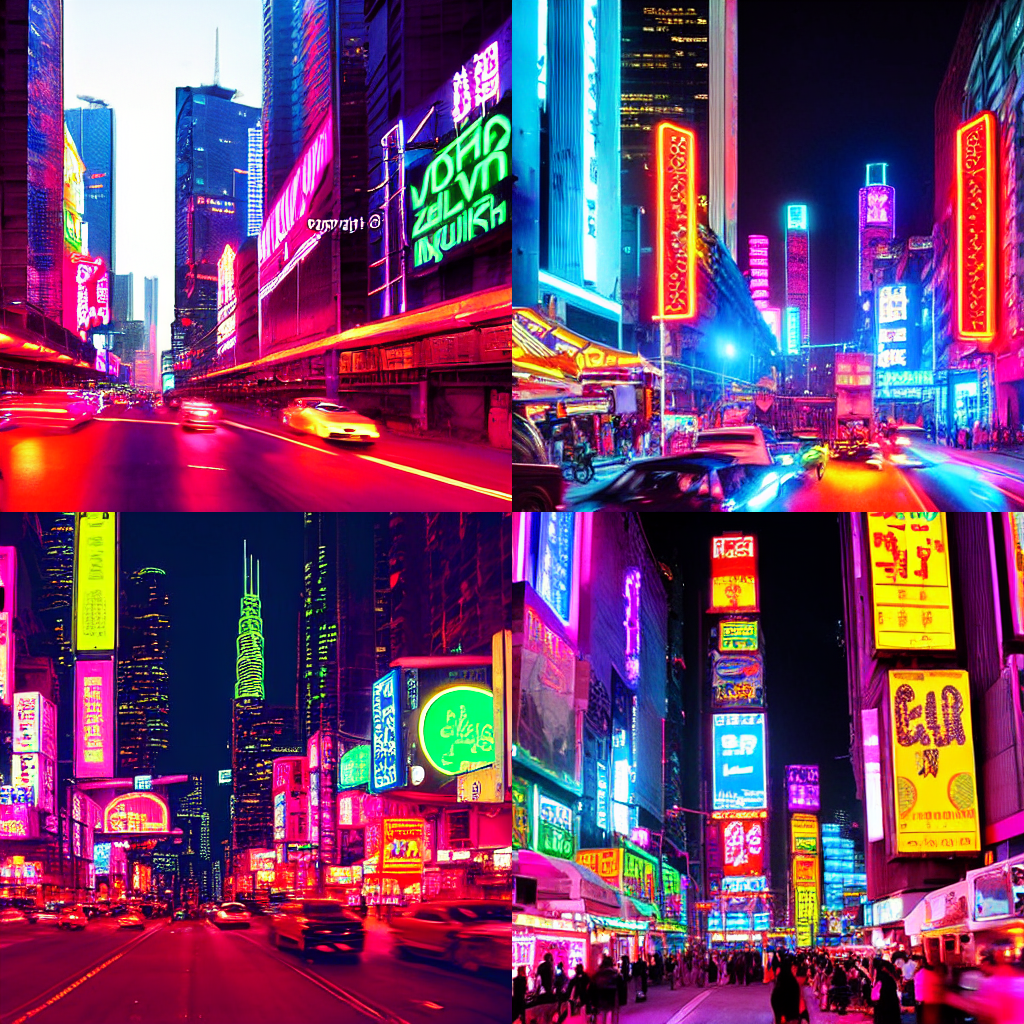} &
  \includegraphics[width=0.3\linewidth]{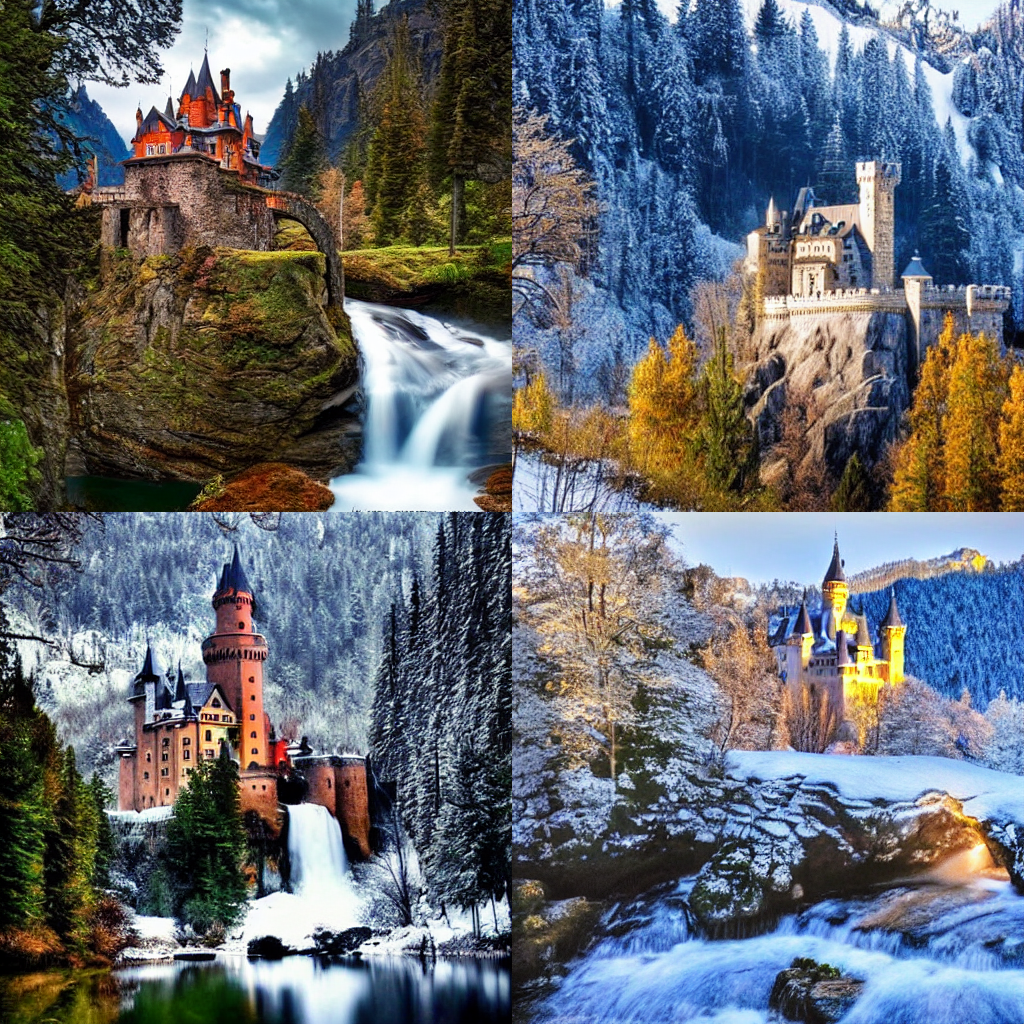} \\
  Cabin in a Snowy Forest & Cyberpunk City & Fairy Tale Castle  \\
\end{tabular}
\caption{Generation results for GPTDrawer across different scenes}
\label{tab:images}
\end{table*}

\begin{abstract}
    In the burgeoning field of AI-driven image generation, the quest for precision and relevance in response to textual prompts remains paramount. This paper introduces GPTDrawer, an innovative pipeline that leverages the generative prowess of GPT-based models to enhance the visual synthesis process. Our methodology employs a novel algorithm that iteratively refines input prompts using keyword extraction, semantic analysis, and image-text congruence evaluation. By integrating ChatGPT for natural language processing and Stable Diffusion for image generation, GPTDrawer produces a batch of images that undergo successive refinement cycles, guided by cosine similarity metrics until a threshold of semantic alignment is attained. The results demonstrate a marked improvement in the fidelity of images generated in accordance with user-defined prompts, showcasing the system's ability to interpret and visualize complex semantic constructs. The implications of this work extend to various applications, from creative arts to design automation, setting a new benchmark for AI-assisted creative processes. 
\end{abstract}    

\section{Introduction}

Stable Diffusion (SD), as delineated by Rombach et al. \cite{rombach2021highresolution}, represents a paradigm shift in the domain of latent diffusion models. It's premised on the ability to transmute text-based prompts into high-resolution visual constructs. These prompts act not just as triggers but as foundational descriptors delineating the contours of the visual content the model generates.

Recent advancements in text-to-image models have been significant, as highlighted in various studies \cite{SD_Survey}. Prominent among these are Dall-E, Imagen, and Midjourney. However, Stable Diffusion has garnered considerable attention, particularly due to its open-source community support. Notable contributions such as ControlNet \cite{ControlNet} and LoRA \cite{LoRA} have provided built-in extensions to Stable Diffusion, enhancing its user-friendliness and accessibility.

However, our preliminary experiments have highlighted a potential limitation in the SD model. As the complexity and granularity of textual descriptions escalate, the model occasionally manifests a decrement in its efficacy, failing to produce visualizations that congruently mirror the descriptive intricacies of the provided text.

\begin{figure}
    \centering
    \includegraphics[width=1\linewidth]{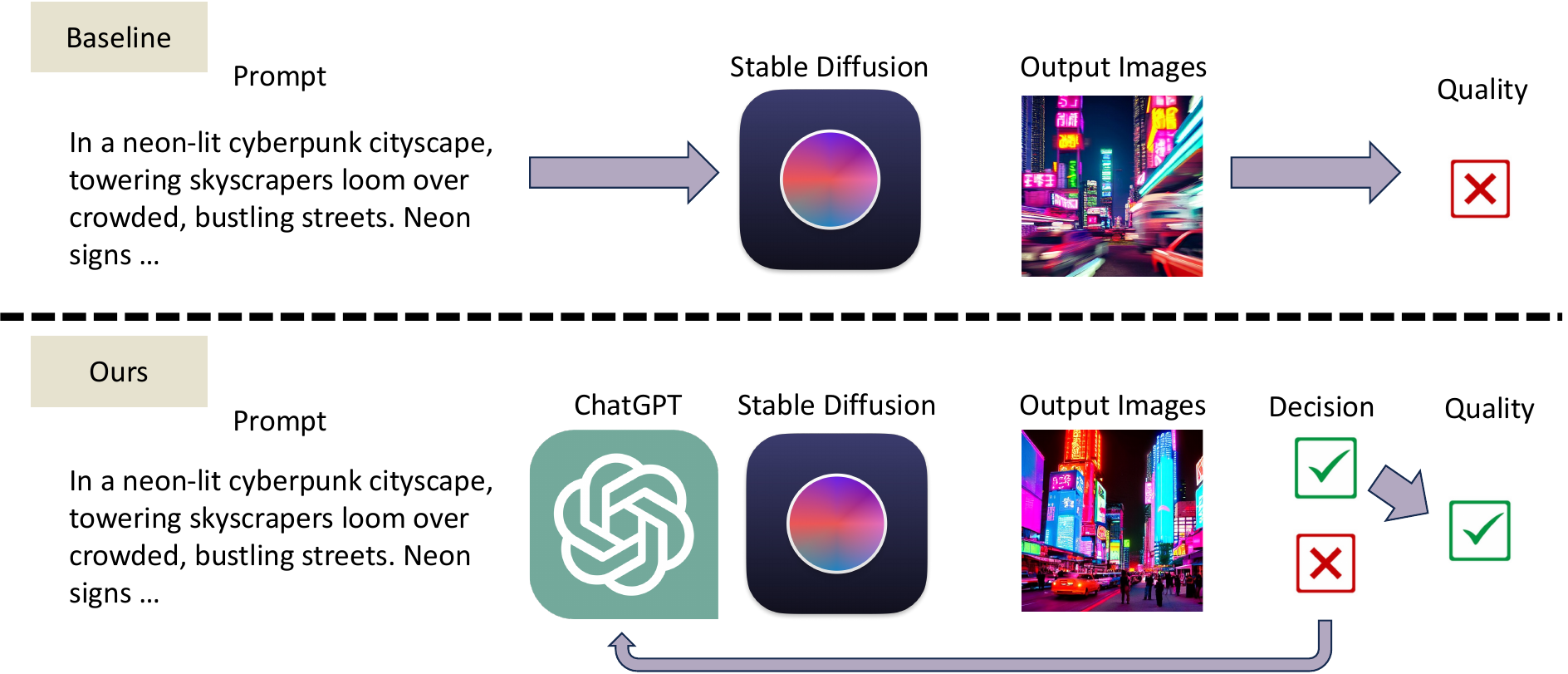}
    \caption{Comparision between original Stable Diffusion AI-generation pipeline with ours GPTDrawer enhancing AI-generation pipeline. In our pipeline, the results have a high possibility of matching with the original provided prompt.} 
    \label{fig/framework}
\end{figure}

Simultaneously, there's been a surge in advancements in the domain of natural language processing, epitomized by models such as ChatGPT. Its prowess in comprehending, interpreting, and generating refined text offers a tantalizing proposition: Can ChatGPT be harnessed as an auxiliary tool, a "prompt architect", to enhance the congruence between intricate textual prompts and the visual outputs of the Stable Diffusion model?

In the ensuing discourse, this report seeks to elucidate the potential of amalgamating ChatGPT's linguistic capabilities with the visual generation potential of SD. To be specific, We introduce 'GPTDrawer', a pipeline designed to create images matching complex textual descriptions of scenes. ChatGPT serves as an intermediary to refine textual prompts, optimizing them for SD's processing framework. To evaluate image quality, we utilize the Vision-Language model BLIP \cite{BLIP} for quantitative similarity assessments. We iteratively refine the prompt for SD if any keywords fail to pass the similarity check of the BLIP. The methodology is detailed in Section 3.  Our experiments, presented in Section 4, demonstrate GPTDrawer's superior performance over the baseline in both qualitative and quantitative measures. Table \ref{tab:images} shows some generation results of our GPTDrawer.

\section{Related Works}

In this section, we discuss the related works of current prompt enhancement methods (Sec.~\ref{sec:rw:prompt}) and Generative Model (Sec.~\ref{sec:rw:generativemodel}).

\subsection{Prompt Enhancement}
\label{sec:rw:prompt}

The concept of prompt enhancement in the field of artificial intelligence has garnered significant attention, particularly in improving the interaction between humans and AI systems~\cite{liang2023iterative, wang2023plate, li2022spe}. This area of study focuses on refining the input prompts provided to AI models to enhance the accuracy, relevance, and creativity of the generated outputs. In the context of language models and text-to-image synthesis, the precision and contextual richness of the input prompts are crucial determinants of the quality and applicability of the outputs.
The integration of AI in creative processes, like text-to-image synthesis, has opened new avenues for prompt enhancement research. The challenge lies in translating abstract or subjective textual descriptions into visually coherent images. Studies such as Radford \textit{et.al.} have made strides in this area, proposing algorithms that effectively bridge the gap between textual input and visual output~\cite{radford2021learning}. These algorithms often rely on intricate processes of keyword extraction, context analysis, and iterative refinement to ensure that the generated images align closely with the user's intent.

\subsection{Generative Model}
\label{sec:rw:generativemodel}

Generative models represent a cornerstone of contemporary AI research, particularly in the realm of automated content creation~\cite{gui2021review, goodfellow2020generative, goodfellow2014generative, yang2023diffusion, rombach2022high, chan2022efficient, kingma2013auto, lopez2018information, van2017neural}. These models are designed to produce new data instances that are indistinguishable from real data, spanning domains from text generation to image synthesis. 
The landscape of generative models witnessed a paradigm shift with the introduction of models like GPT-3 by OpenAI and Google's BERT for natural language understanding and generation~\cite{devlin2018bert, floridi2020gpt, lee2020patent}. These models, based on transformer architectures, have redefined the benchmarks for text generation in terms of fluency, coherence, and context awareness. Their applications extend beyond mere text generation, impacting areas such as conversational AI, content creation, and even coding.
In the realm of image synthesis, the emergence of models like DALL-E and Stable Diffusion marks a new era~\cite{SD_Survey, rombach2021highresolution}. These models, capable of generating high-fidelity images from textual descriptions, have opened up unprecedented possibilities in the field of creative arts and beyond. The research by Rombach demonstrates the potential of these models in artistic creation~\cite{rombach2021highresolution}.

\section{Methodology}

In this section, we discuss our method to refine the Stable Diffusion.

\begin{figure*}
    \centering
    \includegraphics[width=1\textwidth]{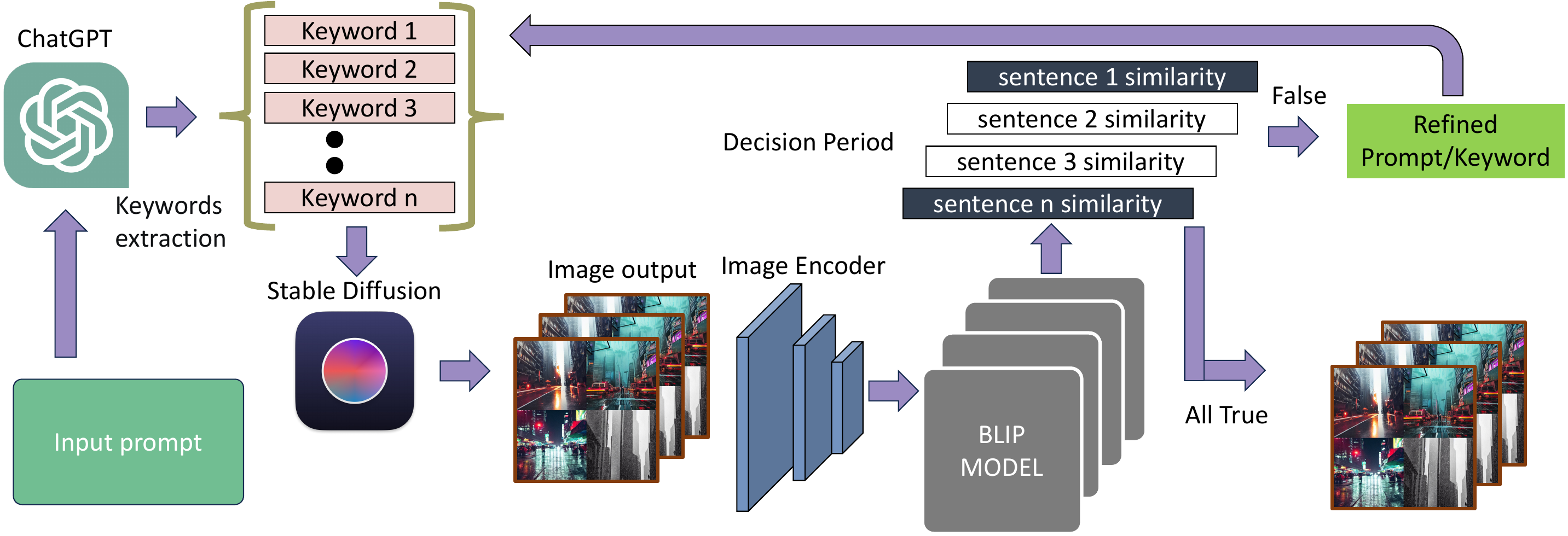}
    \caption{Framework Figure} 
    \label{fig/framework}
\end{figure*}

\subsection{Overview}
\label{sec:overview}
In the proposed workflow delineated in Algorithm~\ref{algorithm:pipeline} and Figure~\ref{fig/framework}, the system commences by ingesting an input prompt, denoted as \textit{P}. This input is subject to a rigorous extraction procedure executed by the language model, \textit{ChatGPT}, which distills an array of keywords, symbolized as $\mathbf{w}$. These keywords serve as the cornerstone for the \textit{Stable Diffusion} algorithm, which synthesizes a batch of preliminary visual representations, designated as $I_{B}$.

Each visual output within the batch $I_{B}$ is subsequently encoded through an Image Encoder, expressed as $E_{image}(I_{B})$. Concurrently, the keyword array $\mathbf{w}$ undergoes a similar encoding process to yield $E_{text}(\mathbf{w})$. The crux of the process involves computing a cosine similarity measure, $Sim_{cos}$, to quantify the alignment between the encoded representations of the visual outputs and the keywords.

The iterative refinement cycle is triggered if the similarity metric for any keyword fails to surpass a pre-established threshold, $T$. During this phase, \textit{ChatGPT} is tasked with refining both the prompt \textit{P} and the individual keywords $\mathbf{w}_{i}$. This refinement is informed by the discrepancies identified through the similarity assessment. After prompt adjustment, the \textit{Stable Diffusion} model generates a new set of images, $I_{B}^{'}$, which are then re-assessed for conformity to the refined keywords.

This recursive process is iterated until the similarity for all keywords meets or exceeds the threshold, at which point the process concludes. The culmination of this methodology is the generation of a set of refined images $I_B$, each associated with a maximal cosine similarity measure, $Sim_{cos}$. The outputs are thus ensured to be in close adherence to the semantic intent encapsulated within the original input prompt.

\subsection{Input Prompt Reception}

At the commencement of the process, the system stands in a state of readiness to receive an input prompt from the user. This stage is critical as it defines the initial condition from which the entire generation process unfolds. We can represent the initial prompt as a variable, denoted as $P_0$, which encapsulates the user's input in a textual format. Formally, the input prompt reception can be described as follows:

\begin{equation}
    P_0 = UserInput
\end{equation}

Here, $P_0$ signifies the initial prompt provided by the user, serving as the foundational input for the subsequent stages of the pipeline. The function \(\text{UserInput}()\) symbolizes the act of receiving input from the user, encapsulating the user's intent, context, and requirements in a format interpretable by the system. This initial prompt is then processed through the pipeline to extract relevant keywords and semantic meanings, setting the stage for the generation of the initial set of images.

\begin{algorithm}
\caption{Image Generation Process workflow}
\begin{algorithmic}[1]

\State Input prompt \textit{P}
\State Extract keywords array $\mathbf{w}$ with \textit{ChatGPT}.
\State Generate with \textit{Stable Diffusion} for $B$ images as $I_{B}$.
\Loop
    \State Encode image with Image Encoder $E_{image}(I_{B})$
    \State Encode keywords array $\mathbf{w}$ as $E_{text}(\mathbf{w})$.
    \State Calculate $Sim_{cos}$ with image and sentence.
    \If {for all $E_{text}(\mathbf{w})$ larger than threshold $T$}
        \State \textbf{break}
    \Else
        \State Refine prompt \textit{P} and words $\mathbf{w}_{i}$ with \textit{ChatGPT}
        \State Generate $I_{B}^{'}$ with Stable Diffusion.
        \State Assign $I_{B}^{'}$ to $I_{B}$.
    \EndIf
\EndLoop
\State Output final refined images $I_B$ and maximal similarity $Sim_{cos}$

\end{algorithmic}
\label{algorithm:pipeline}
\end{algorithm}

\subsection{Keyword Extraction}

Upon receiving the initial prompt \( P_0 \), the system proceeds to the keyword extraction phase. This phase aims to distill the most essential and semantically significant terms from the input. Utilizing advanced natural language processing (NLP) techniques, the system analyzes \( P_0 \) and segments it into a set of key terms, denoted as \( W \).

The extraction process can be mathematically represented as:

\begin{equation}
    W = ExtractKeywords (P_0)
\end{equation}

Here, \( ExtractKeywords(\cdot) \) is a function embodying the ChatGPT algorithms used for keyword extraction. 
Simultaneously, the system filters out stop words and other irrelevant terms to ensure the purity and relevance of the extracted keywords \( W \). This filtration process enhances the precision of the subsequent image generation stages, ensuring that the generated content is closely aligned with the user's original intent encapsulated in \( P_0 \).

\subsection{Image Generation}
\label{sec:imagegeneration}

Following the extraction of keywords $W$ from the initial prompt $P_0$, the subsequent critical phase in the pipeline is the generation of images. This stage entails the transformation of the textual descriptors encapsulated in $W$ into their corresponding visual representations. A generative model, such as \textit{Stable Diffusion}, is employed to synthesize images grounded in the textual descriptions.

The image generation process can be mathematically represented as:

$$ I = \text{StableDiffusion}(W) $$

Here, $I$ denotes the set of generated images, and $\text{StableDiffusion}(\cdot)$ symbolizes the operation of the generative model. This function ingests the extracted keywords $W$ and outputs a collection of initial images. Each image within this collection, $I$, is then subject to an evaluation to gauge its relevance and conformity to the semantic nuances of $W$.

\subsection{BLIP Model Evaluation}

\subsection{BLIP Evaluation}
\label{sec:blip-evaluation}

The integration of the BLIP (Bootstrapped Language Image Pretraining) model within our image generation framework plays a pivotal role in ensuring semantic congruence between the generated images and the textual prompts. This subsection elucidates the operational specifics and the evaluative function of BLIP in our pipeline.

\paragraph{Operational Methodology}
The process initiates with the loading of the generated image, employing the \texttt{load\_image} function, which is meticulously calibrated to adhere to a pre-set \texttt{image\_size} and tailored for a specific computational \texttt{device}. Following this, the BLIP model, pretrained on the COCO dataset and adapted for the said \texttt{image\_size}, is instantiated and configured to the evaluation mode.

\paragraph{Textual Input Processing}
In our setup, the BLIP model processes a detailed descriptive sentence alongside a set of concisely curated keywords. These textual inputs represent diverse facets of the intended visual output, ranging from overarching themes to specific elements.

\paragraph{Cosine Similarity Computation}
The BLIP model's evaluative mechanism is centered around the calculation of cosine similarity between the image and text feature vectors. This is mathematically represented as:
\begin{equation}
    \text{Cosine Similarity}(\vec{E}_{\text{img}}, \vec{E}_{\text{text}}) = \frac{\vec{E}_{\text{img}} \cdot \vec{E}_{\text{text}}}{\|\vec{E}_{\text{img}}\| \times \|\vec{E}_{\text{text}}\|}
\end{equation}
where $\vec{E}_{\text{img}}$ and $\vec{E}_{\text{text}}$ are the features of encoding image and text, respectively. This metric quantifies the alignment of the image with the individual textual elements, as well as with the comprehensive narrative description.

\subsection{Prompt Refinement}
\label{sec:prompt-refinement}

The efficacy of the image generation process in our pipeline is significantly influenced by the precision and relevance of the input prompts. In instances where the cosine similarity score, as evaluated by the BLIP model, is found to be below a predefined \textit{threshold} $T$, indicating a low degree of semantic alignment between the generated image and the textual prompt, we initiate a prompt refinement process.

\paragraph{Methodology for Refinement}
The refinement strategy is predicated on the adaptation of the original keywords to more general versions. This approach is rooted in the hypothesis that broader keywords may encapsulate a wider range of visual interpretations, thereby increasing the likelihood of generating an image that aligns more closely with the user's intent. The process involves:

\begin{itemize}
    \item Analyzing the specific keywords that contributed to the low cosine similarity score.
    \item Substituting these keywords with their more generalized counterparts.
    \item Ensuring that the modified prompt retains the essential thematic elements of the original prompt while broadening its interpretative scope.
\end{itemize}

The prompt refinement is inherently iterative. Following the modification of the prompt as we mentioned in algorithm~\ref{algorithm:pipeline}:

\begin{enumerate}
    \item The revised keywords is fed back into the image generation model.
    \item A new set of images is generated based on the updated prompt.
    \item These images are then re-evaluated for their cosine similarity with the refined prompt.
\end{enumerate}

\section{Experiments}

We evaluate the performance of our proposed GPTDrawer over three scenes with detailed descriptions. 


\textbf{Scene I: CyberPunk City} In a neon-lit cyberpunk cityscape, towering skyscrapers loom over crowded, bustling streets. Neon signs in various languages flicker, casting a colorful glow on the eclectic mix of pedestrians and street vendors. Cars zip by above, while a large digital billboard displays futuristic advertisements.

\textbf{Scene II: Fairy Tale Castle} High in a fairy-tale realm, a majestic mountain towers, crowned with glittering snow. Cascading waterfalls and crystal-clear streams flow from its heights, nourishing lush valleys below. At its peak, an ancient castle of stone and enchantment stands, overlooking a realm of wonder.

In our GPTDrawer framework, we set the cosine similarity score threshold $T$ at 0.2, as detailed in Sections 3.5 and 3.6. If the cosine similarity of any keyword falls below this threshold, we revise the prompt or increase the keyword's weight before initiating another generation cycle. Each generation cycle produces a batch of 16 images.

We use the results generated by inputting detailed descriptions directly into Stable Diffusion models as our baseline. Each scene undergoes both qualitative and quantitative evaluations. For the qualitative evaluation, we manually assess whether the generation results align with each keyword identified by ChatGPT. For the quantitative evaluation, we employ BLIP to measure the similarity between the generated images and each keyword or sentence.





\subsection{Result on Scene I}
\textbf{Keywords extracted by ChatGPT:} Cyberpunk cityscape, Neon-lit skyscrapers, Crowded streets, Multilingual neon signs, Colorful glow, Eclectic pedestrians, Street vendors, Cars, Futuristic digital billboard, Advertisements.

Table \ref{tab:images_2} displays the generation outcomes for Scene I using GPTDrawer. Initially, the generated result failed to incorporate the keyword 'cars'. To address this, we increased the weight of 'cars' to 1.1 in the Stable Diffusion Model. This adjustment led to successful results in the second generation round, satisfying all keyword similarity criteria, with details available in Table \ref{tab:keywords}. In contrast, the baseline results failed to depict 'Eclectic pedestrians' and 'Street vendors', missing these crucial keywords. Conversely, GPTDrawer's final output adhered to all keyword requirements.

\begin{table}[h]
\centering
\begin{tabular}{ccc}
  \includegraphics[width=0.3\linewidth]{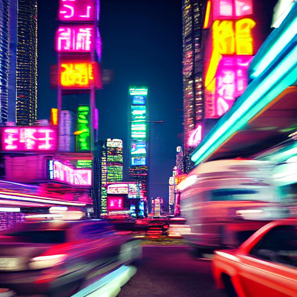} & 
  \includegraphics[width=0.3\linewidth]{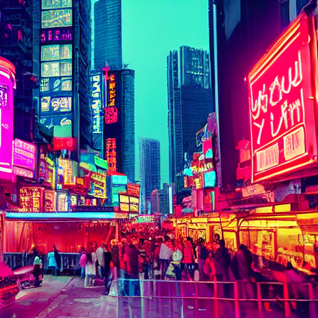} &
  \includegraphics[width=0.3\linewidth]{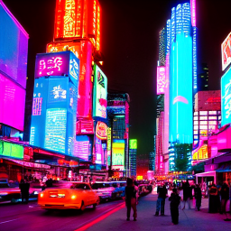} \\
  Baseline & GPTDrawer-1st& GPTDrawer-final
\end{tabular}
\caption{Generation results for Scene I}
\label{tab:images_2}
\end{table}

Table \ref{tab:sentences} provides the sentence similarity evaluation. GPTDrawer excels in the overall ratings for Scene I, affirming its efficacy in producing images that more accurately reflect the specified description.

\subsection{Result on Scene II}
\textbf{Keywords extracted by ChatGPT:} Fairy-tale realm, Mountain, Snow, Waterfalls, Streams, Valleys, Castle, Enchantment, Wonder.
\begin{table}[h]
\centering
\begin{tabular}{cc}
  \includegraphics[width=0.45\linewidth]{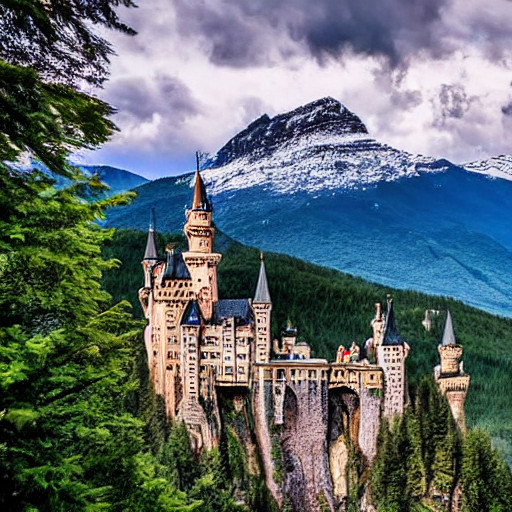} & 
  \includegraphics[width=0.45\linewidth]{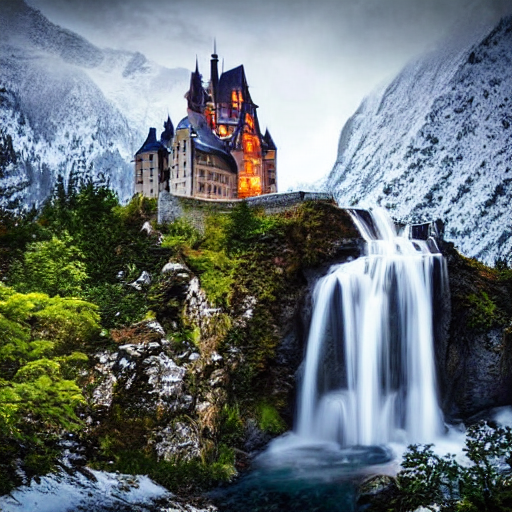} \\
  Baseline & GPTDrawer \\
\end{tabular}
\caption{Generation results for Scene II}
\label{tab:images_3}
\end{table}

\begin{table*}[]
\centering
\begin{tabular}{|m{2cm}|m{3.5cm}|c|c|c|c|}
\hline
& & \multicolumn{2}{c|}{Human Eye Perception} & \multicolumn{2}{c|}{BLIP keyword Cosine Similarity} \\ \cline{3-6}
\multirow{-2}{*}{Scenes} & \multirow{-2}{*}{Keywords} & Baseline & GPTDrawer & Baseline & GPTDrawer \\ \hline
\multirow{8}{*}{Scene I} & Cozy, rustic cabin & \cellcolor{red}False & True & \cellcolor{red}0.1483 & 0.3716 \\ \cline{2-6}
 & Snowy forest & True & True & 0.3934 & 0.4124 \\ \cline{2-6}
 & Twilight & True & True & 0.2260 & 0.2153 \\ \cline{2-6}
 & Chimney & \cellcolor{red}False & True & \cellcolor{red}0.0959 & 0.2378 \\ \cline{2-6}
 & Glowing windows & \cellcolor{red}False & True & \cellcolor{red}0.1930 & 0.2868 \\ \cline{2-6}
 & Snow-covered path & True & True & 0.4002 & 0.3494 \\ \cline{2-6}
 & Tall pine trees & True & True & 0.3796 & 0.3798 \\ \cline{2-6}
 & Snow-laden branches & True & True & 0.2834 & 0.3487 \\ \hline
\multirow{8}{*}{Scene II} & Cyberpunk cityscape & True & True & 0.3575 & 0.3678 \\ \cline{2-6}
 & Neon-lit skyscrapers & True & True & 0.4476 & 0.4590 \\ \cline{2-6}
 & Crowded streets & \cellcolor{red}False & True & 0.3590 & 0.3709 \\ \cline{2-6}
 & Multilingual neon signs & True & True & 0.3862 & 0.3730 \\ \cline{2-6}
 & Colorful glow & True & True & 0.3575 & 0.3902 \\ \cline{2-6}
 & Eclectic pedestrians & \cellcolor{red}False & True & \cellcolor{red}0.1927 & 0.2278 \\ \cline{2-6}
 & Street vendors & \cellcolor{red}False & True & \cellcolor{red}0.1984 & 0.2677 \\ \cline{2-6}
 & Cars & True & True & 0.3401 & 0.2846 \\ \cline{2-6}
 & Futuristic billboard & True & True & 0.3455 & 0.3379 \\ \cline{2-6}
 & Advertisements & True & True & 0.2488 & 0.2344 \\ \hline
\multirow{9}{*}{Scene III} & fairy-tale realm & True & True & 0.2918 & 0.2731 \\ \cline{2-6}
 & mountain & True & True & 0.3618 & 0.3228 \\ \cline{2-6}
 & snow & True & True & 0.2627 & 0.2477 \\ \cline{2-6}
 & waterfalls & \cellcolor{red}False & True & \cellcolor{red}0.1886 & 0.3705 \\ \cline{2-6}
 & streams & \cellcolor{red}False & True & \cellcolor{red}0.1841 & 0.2140 \\ \cline{2-6}
 & valleys & \cellcolor{red}False & \cellcolor{red}False & \cellcolor{red}0.1908 & 0.2168 \\ \cline{2-6}
 & castle & True & True & 0.3441 & 0.3444 \\ \cline{2-6}
 & enchantment & True & True & 0.2606 & 0.2263 \\ \cline{2-6}
 & wonder & True & True & 0.3050 & 0.2710 \\ \hline
\end{tabular}
\caption{Qualitative and quantitative keyword checks over different scenes on keywords. Keywords that haven't passed qualitative or quantitative checks are highlighted in red.}
\label{tab:keywords}
\end{table*}
\begin{table*}[h]
\centering
\begin{tabular}{|m{2cm}|m{3cm}|c|c|}
\hline
\textbf{Scenes} & \textbf{Sentences} & \textbf{Baseline} & \textbf{GPTDrawer} \\ \hline
\multirow{5}{*}{Scene I} & Sentence 1 & 0.4103 & \textbf{0.4972} \\ \cline{2-4}
 & Sentence 2 & 0.1644 & \textbf{0.3608} \\ \cline{2-4}
 & Sentence 3 & 0.3870 & \textbf{0.4157} \\ \cline{2-4}
 & Sentence 4 & \textbf{0.4415} & 0.3375 \\ \cline{2-4}
 & Overall & 0.3820 & \textbf{0.4894} \\ \hline
\multirow{4}{*}{Scene II} & Sentence 1 & 0.4347 & \textbf{0.4683} \\ \cline{2-4}
 & Sentence 2 & 0.3761 & \textbf{0.4236} \\ \cline{2-4}
 & Sentence 3 & \textbf{0.4494} & 0.3818 \\ \cline{2-4}
 & Overall & 0.4510 & \textbf{0.4621} \\ \hline
\multirow{4}{*}{Scene III} & Sentence 1 & \textbf{0.4840} & 0.4572 \\ \cline{2-4}
 & Sentence 2 & 0.3773 & \textbf{0.4356} \\ \cline{2-4}
 & Sentence 3 & \textbf{0.4909} & 0.4666 \\ \cline{2-4}
 & Overall & 0.4394 & \textbf{0.5106} \\ \hline
\end{tabular}
\caption{Quantitative similarity assessment over different scenes on sentences. Better similarity scores are highlighted.}
\label{tab:sentences}
\end{table*}

Table \ref{tab:images_3} details the generation results for Scene II using GPTDrawer. In its first generation round, GPTDrawer met all keyword similarity requirements, making this initial result the definitive output, as further detailed in Table \ref{tab:keywords}. The baseline, however, failed to incorporate significant elements such as 'waterfalls', 'streams', and 'valleys'. While GPTDrawer also overlooked the 'valley' keyword according to human eye assessment, it successfully captured all other keywords.

The sentence similarity evaluations are presented in Table \ref{tab:sentences}. Here, GPTDrawer outshines in the overall ratings for Scene II, clearly demonstrating its superior ability to generate images that more closely align with the provided description.

\section{Conclusion}
This report has comprehensively explored the integration of ChatGPT's advanced linguistic processing with Stable Diffusion (SD)'s visual generation prowess, culminating in the development of the innovative 'GPTDrawer' pipeline. Our investigations underscored SD's challenges in accurately translating complex textual prompts into congruent visual outputs. GPTDrawer, however, emerges as a groundbreaking solution, leveraging ChatGPT's capabilities to refine these prompts, thereby enhancing SD's effectiveness.

Our experimental results, as elaborated in Section 4, provide compelling evidence of GPTDrawer's superiority over traditional SD approaches. By employing the Vision-Language model BLIP for quantitative similarity assessments, we established that GPTDrawer not only meets but often surpasses the baseline in both qualitative and quantitative aspects. This is particularly notable in its ability to handle intricate, detailed prompts, where standard SD methods showed limitations.

Furthermore, the success of GPTDrawer illuminates a promising pathway for future research in latent diffusion models and natural language processing. It exemplifies the potential of symbiotic AI systems where complementary technologies are harnessed to overcome individual limitations, setting a new benchmark in the realm of AI-driven image generation.
{
    \small
    \bibliographystyle{ieeenat_fullname}
    \bibliography{report}
}


\end{document}